%% file: acl_latex.tex
\pdfoutput=1

\documentclass[11pt]{article}

\usepackage[final]{acl}

\usepackage{times}
\usepackage{latexsym}
\usepackage{graphicx}
\usepackage{booktabs}
\usepackage{tcolorbox}
\usepackage{hyperref}
\usepackage{dblfloatfix}

\usepackage[T1]{fontenc}

\usepackage[utf8]{inputenc}

\usepackage{microtype}
\usepackage{bm}
\usepackage{mathtools}
\usepackage{amsmath,amsfonts,amssymb,amsthm}
\usepackage[ruled,noend]{algorithm2e}
\usepackage{multirow}
\usepackage{enumitem}
\usepackage{soul} 

\usepackage{cleveref}
\crefformat{section}{\S#2#1#3}
\crefformat{subsection}{\S#2#1#3}
\crefformat{subsubsection}{\S#2#1#3}
\crefrangeformat{section}{\S\S#3#1#4 to~#5#2#6}
\crefmultiformat{section}{\S\S#2#1#3}{ and~#2#1#3}{, #2#1#3}{ and~#2#1#3}
\usepackage{refstyle}
\Crefformat{figure}{#2Figure~#1#3}
\Crefmultiformat{figure}{Figures~#2#1#3}{ and~#2#1#3}{, #2#1#3}{ and~#2#1#3}
\Crefformat{table}{#2Table~#1#3}
\Crefmultiformat{table}{Tables.~#2#1#3}{ and~#2#1#3}{, #2#1#3}{ and~#2#1#3}
\Crefformat{appendix}{#2Appx.~\S#1#3}
\crefformat{algorithm}{Alg.~#2#1#3}
\Crefformat{equation}{#2Eq.~#1#3}

\definecolor{darkgreen}{HTML}{008f00}
\definecolor{UMDred}{HTML}{ed1c24}
\newcommand{\roberta}{\textsc{RoBERTa}}
\newcommand{\bert}{\textsc{BERT}}
\newcommand{\albert}{\textsc{ALBERT}}
\newcommand{\palm}{\textsc{PaLM}}
\newcommand{\gpttwo}{\textsc{GPT-2}}
\newcommand{\gptthree}{\textsc{GPT-3}}

\title{Do ever larger octopi still amplify reporting biases?\\ Evidence from judgments of typical colour}

\author{Fangyu Liu$^{1}$, Julian Martin Eisenschlos$^2$, Jeremy R. Cole$^2$, Nigel Collier$^1$ \\
$^1$ University of Cambridge \ \ \ \ \  $^2$ Google Research \\
\texttt{\{fl399, nhc30\}@cam.ac.uk \ \ \{eisenjulian,jrcole\}@google.com}
}

\begin{document}
\maketitle
\begin{abstract}
Language models (LMs) trained on raw texts have no direct access to the physical world. \citet{gordon2013reporting} point out that LMs can thus suffer from \textit{reporting bias}: texts rarely report on common facts, instead focusing on the unusual aspects of a situation. If LMs are only trained on text corpora and naively memorise local co-occurrence statistics, they thus naturally would learn a biased view of the physical world. While prior studies have repeatedly verified that LMs of smaller scales (e.g., \roberta, \gpttwo) amplify reporting bias, it remains unknown whether such trends continue when models are scaled up. We investigate reporting bias from the perspective of colour in larger language models (LLMs) such as \palm{} and \gptthree. Specifically, we query LLMs for the typical colour of objects, which is one simple type of perceptually grounded physical common sense. Surprisingly, we find that LLMs significantly outperform smaller LMs in determining an object's typical colour and more closely track human judgments, instead of overfitting to surface patterns stored in texts.
This suggests that very large models of language alone are able to overcome certain types of reporting bias that are characterized by local co-occurrences.\footnote{\href{https://github.com/google-research/language/tree/master/language/octopus-llm}{https://github.com/google-research/language/tree/master/language/octopus-llm} (code).}
\end{abstract}

\input{tex/introduction}
\input{tex/methods}

\input{tex/results}

\input{tex/limitations}
\input{tex/conclusion}

\input{tex/ack}

\bibliography{anthology,custom}
\bibliographystyle{acl_natbib}

\appendix

\clearpage 

\begin{table*}[!t]
    \centering
    \scalebox{0.78}{
    \begin{tabular}{lcccccccccc}
    \toprule
    & & \multicolumn{3}{c}{0-shot} &  \multicolumn{3}{c}{1-shot} & \multicolumn{3}{c}{5-shot}   \\
    \cmidrule(lr){3-5}     \cmidrule(lr){6-8}     \cmidrule(lr){9-11}
    Model & Size & Acc$_{@1}$ & $\rho_{\text{human}}$ & $\rho_{\text{ngram}}$  &  Acc$_{@1}$ & $\rho_{\text{human}}$ & $\rho_{\text{ngram}}$ &  Acc$_{@1}$ & $\rho_{\text{human}}$ & $\rho_{\text{ngram}}$ \\
    \midrule
    Google Ngram & - &  36.3 & 44.2 & 100.0 & - & - & - & - & - & - \\
    \midrule 
    \roberta$_{\text{Large}}$$^*$ & 335M & 37.6 & - & - & - & - & - & - & - & - \\
     \roberta$_{\text{Large}}$ (real) & 335M & 7.3 & 25.8 & 55.9 & 8.4 & 17.0 & 52.9 & 15.4 & 28.5 & 51.8 \\
    \midrule
    \gptthree$_{\text{davinci}}$ & 175B & \textbf{55.5} & \textbf{43.1} & 65.0 & 61.8 & 60.5  & 61.0 & 63.1 & 62.3 & 55.9 \\
    \palm-540B & 540B & 42.6 & 46.0 & 66.3 & \textbf{63.9} & \textbf{62.5} & 62.5 & \textbf{64.9} & \textbf{66.2} & 60.1\\
    \bottomrule
    \end{tabular}
    }
    \caption{Evaluating the best performing SLM on CoDa, using one consistent prompt (the same setup for all LLMs tested). Performance of the optimistic 10-prompt setup by \citet{paik-etal-2021-world} and also performance of LLMs are listed for reference. When evaluated under the same protocol as LLMs, the best performing SLM \roberta's performance drops very significantly and is at chancel level.}
    \label{tab:fair_setup}
\end{table*}

\section{A More Comprehensive Table (\Cref{tab:main_results_full})}\label{sec:comprehensive}
In the main text, we compare different models under different setups in \Cref{tab:main_results}. To provide more information for reference and also strengthen our findings, we present a more comprehensive \Cref{tab:main_results_full} which also reports Kendall's $\tau$ as a correlation metric, and include Wikipedia stats provided by \citet{paik-etal-2021-world} as another source of ngrams. The main conclusion remains the same. Kendall's $\tau$ has identical trend to Spearman's $\rho$, and similar fitting trend of Google Ngram is also shown on Wikipedia.

\begin{table*}[!t]
    \centering
    \scalebox{0.86}{
    \begin{tabular}{lcccccccc}
    \toprule
    & & \multicolumn{3}{c}{human} & \multicolumn{2}{c}{GBN} & \multicolumn{2}{c}{wiki} \\
   \cmidrule(lr){3-5} \cmidrule(lr){6-7} \cmidrule(lr){8-9}
    Model & size & Acc$_{@1}$ & $\rho$ & $\tau$ & $\rho$ & $\tau$ & $\rho$ & $\tau$ \\
    \midrule
    GBN & - &  36.3 & 44.2 & 36.2 & 100.0 & 100.0 & 66.5 & 55.9 \\
    wiki & - & 23.3	& 28.6 & 23.2  & 66.5	& 55.9  & 100.0 & 100.0 \\
    \midrule 
    \multicolumn{9}{c}{\textit{0-shot}} \\
    \midrule
    \roberta$_{\text{Base}}$$^*$ & 110M & 28.0 & - & - & - & - & - & -\\
    \roberta$_{\text{Large}}$$^*$ & 335M & 37.6 & - & - & - & - & - & -\\
    \midrule
    \gpttwo$_{\text{Small}}$$^*$ & 124 & 27.1  & - & - &  - & - & - & - \\
    \gpttwo$_{\text{Base}}$$^*$ & 355M & 31.7 & - & - &  - & - & - & - \\
    \gpttwo$_{\text{Large}}$$^*$ & 774M & 33.2 & - & - &  - & - & - & - \\
    \gpttwo$_{\text{XL}}$$^*$ & 1.5B & 36.1 & - & - &  - & - & - & - \\
    \midrule
    \albert$_{\text{V2-Base}}$$^*$ & 11M & 20.9 & - & - & - & - & - & - \\
    \albert$_{\text{V2-Large}}$$^*$ & 17M & 28.8 & - & - & - & - & - & - \\
    \albert$_{\text{V2-XL}}$$^*$ & 58M  & 25.2 & - & - & - & - & - & - \\
    \albert$_{\text{V2-XXL}}$$^*$ & 223M & 31.8 & - & - & - & - & - & - \\
    \midrule
    T5$_{\text{Large}}$ & 770M & 21.1 & 25.7 & 20.6 & 42.2 & 32.1 & 33.3 & 25.9 \\
    T5$_{\text{XL}}$ & 3B & 44.3 & 57.4 & 46.6 & 60.3 & 47.3 & 41.7 & 32.3 \\
    T5$_{\text{XXL}}$ & 11B & 50.9 & 49.5 & 40.5 & 57.5 & 44.9 & 40.5 & 31.4   \\
    \midrule
    \gptthree$_{\text{ada}}$ & 350M  & 17.9	& 20.3 & 15.7 & 48.8 & 36.7 & 36.9 & 28.1 \\
    \gptthree$_{\text{babbage}}$ & 1.3B & 27.6 & 27.8 & 22.1 & 58.0 & 44.5 & 44.6 & 34.6 \\
    \gptthree$_{\text{curie}}$ & 6.7B & 33.6 & 41.0 & 32.8 & 63.5 & 50.1 & 37.3 & 36.8 \\
    \gptthree$_{\text{davinci}}$ & 175B & \textbf{55.5} & \textbf{52.8} & \textbf{43.1} & 65.0 & 51.5 & 48.1 & 37.3   \\
    \midrule 
    \palm-8B & 8B & 29.6 & 34.7 & 27.3 & 61.5 & 47.6 & 46.8 & 36.5  \\
    \palm-62B & 62B & 34.2 & 33.5 & 26.9 & 64.4 & 50.9 & 49.9 & 49.5 \\
    \palm-540B & 540B & 42.6 & 44.0 & 35.5 & \textit{66.3} & \textit{52.7} & \textit{48.3} & \textit{38.0}  \\
\midrule 
\multicolumn{9}{c}{\textit{1-shot}} \\
\midrule 
    T5$_{\text{Large}}$ & 770M & 19.4 & 21.0 & 16.4 & 20.3 & 15.7 & 24.5 & 18.5  \\
    T5$_{\text{XL}}$ & 3B & 39.0 & 48.8 & 39.4 & 37.6 & 28.9 & 55.2 & 42.6 \\
    T5$_{\text{XXL}}$ & 11B & 47.2 & 54.3 & 44.3 & 38.7 & 29.6 & 55.9 & 43.5 \\
    \midrule
    \gptthree$_{\text{ada}}$ & 350M  & 21.3 & 24.5 & 19.3 & 46.0 & 35.0 & 34.8 & 27.0 \\
    \gptthree$_{\text{babbage}}$ & 1.3B & 27.6 & 29.8 & 23.6 & 51.7 & 39.7 & 39.7 & 30.5 \\
    \gptthree$_{\text{curie}}$ & 6.7B & 40.1 & 44.2 & 35.6 & 59.2 & 46.3 & 44.6 & 34.7 \\
    \gptthree$_{\text{davinci}}$ & 175B &  61.8 & 60.5 & 50.1 & 61.0 & 48.0 & 42.0 & 32.7    \\
    \midrule 
    \palm-8B & 8B & 39.9 & 48.0 & 38.9 & 64.7 & \textit{51.7} & \textit{47.6} & \textit{37.6} \\
    \palm-62B & 62B &  50.1 & 54.9 & 44.8 & \textit{65.3} & \textit{51.7} & 46.2 & 35.8 \\
    \palm-540B & 540B &  \textbf{63.9} & \textbf{63.5} & \textbf{52.8} & 62.5 & 49.3 & 42.7 & 33.1  \\
\midrule 
\multicolumn{9}{c}{\textit{5-shot}} \\
\midrule 
    T5$_{\text{Large}}$ & 770M &  17.9 & 20.7 & 16.2 & 11.8 & 9.1 & 6.0 & 4.3 \\
    T5$_{\text{XL}}$ & 3B & 42.4 & 47.8 & 38.8 & 60.3 & 47.3 & 42.6 & 33.3 \\
    T5$_{\text{XXL}}$ & 11B & 48.0 & 53.4 & 43.6 & 54.1 & 42.0 & 36.6 & 28.8 \\
    \midrule
    \gptthree$_{\text{ada}}$ & 350M  &  20.5 & 25.4 & 19.9 & 42.2 & 32.3 & 31.2 & 23.8 \\
    \gptthree$_{\text{babbage}}$ & 1.3B & 28.8 & 37.1 & 29.5 & 51.9 & 39.7 & 39.6 & 30.6 \\
    \gptthree$_{\text{curie}}$ & 6.7B & 42.4 & 47.1 & 38.0 & 57.1 & 44.8 & 40.9 & 32.1 \\
    \gptthree$_{\text{davinci}}$ & 175B & 63.1 & 62.3 & 51.6 & 55.9 & 43.7 & 35.9 & 27.7 \\
    \midrule 
    \palm-8B & 8B & 43.8 & 52.3 & 42.6 & \textit{62.0} & \textit{49.1} & \textit{44.9} & \textit{35.1} \\
    \palm-62B & 62B &  58.2 & 61.9 & 51.2 & 61.1 & 48.0 & 41.3 & 31.8  \\
    \palm-540B & 540B &  \textbf{64.9} & \textbf{66.2} & \textbf{55.2} & 60.1 & 47.3 & 40.7 & 31.6  \\    
    \bottomrule
    \end{tabular}
    }
    \caption{Full table containing more corpus stats (wiki) and more metrics (Kendall's $\tau$). GBN: Google Ngram;  wiki: Wikipedia ngrams. Both are from \citet{paik-etal-2021-world}.}
    \label{tab:main_results_full}
\end{table*}

\section{Further Discussions}
\label{sec:further_discussions}

Here we present some more extensive discussions on several topics that concern the experimental setup, including testing SLMs under the same setup as the LLMs (\Cref{sec:fair_setup}); testing different prompts (\Cref{sec:prompt}); the discrepancies among analysed corpora and the real pretraining corpora of LLMs (\Cref{sec:corpora}); the risk of direct data leakage (\Cref{sec:data_leackage}); and error analysis (\Cref{sec:error_analysis}); .

\subsection{Real zero/few-shot setup for SLMs}\label{sec:fair_setup}

In the main text, we used SLM numbers reported by \citet{paik-etal-2021-world} under an optimistic setup: i.e. out of 10 prompts, choosing always the prompt that maximises per-object's performance when evaluating models. We note that when under the same evaluation protocol as LLMs, SLMs' performance would have dropped to chance level. We pick the best performing SLM \roberta$_{\text{Large}}$ as an example. When consistently using one prompt, \roberta$_{\text{Large}}$ has only an accuracy score of 7.3\%. Prompting with few-shot examples does help a bit. However, the 5-shot accuracy of \roberta$_{\text{Large}}$ (real) still has a roughly 50\% gap compared with few-shot performance of the best LLMs.

\subsection{LLMs' Sensitiveness to Prompts}\label{sec:prompt}
For the main experiment, we choose an arbitrary prompt: ``It is known that most \texttt{\{OBJECT\}} have the color \texttt{<mask>}.''. However, it is possible that LLMs are particularly good or bad at this prompt and it is worth testing whether LLMs are robust to how we ask the question. In \Cref{tab:prompt}, we test \gptthree's sensitivity towards different prompts. First, we change the quantifier ``most'' to ``all'', no quantifier, ``some'', ``few'', and ``no''. We find that the LLM is sensitive to the quantifier and produces scores generally well correspond to the quantity being asked. Note that ``all'' and no quantifier lead to lower performance than ``most'', possibly due to the question is unnatural since there is rarely any object exclusively having only one colour. We also paraphrase the original prompt and find that a grammatical paraphrased query can lead to up to around +/-6\% performance difference.  An ungrammatical prompt will damage the model's performance, even including key words such as ``most'', ``color'', and ``common sense''.

\begin{table*}[!h]
    \centering
    \scalebox{0.9}{
    \begin{tabular}{lcc}
    \toprule
    Prompt &  Acc$_{@1}$  \\
    \midrule
It is known that most \texttt{\{OBJECT\}} have the color \texttt{<mask>} \textit{(original)} & 55.5 \\
\midrule
\multicolumn{2}{c}{\textit{different quantifiers}} \\
\midrule
It is known that \underline{all} \texttt{\{OBJECT\}} have the color \texttt{<mask>} & 49.9 \\
It is known that \underline{ }\texttt{\{OBJECT\}} have the color \texttt{<mask>} & 46.3 \\
It is known that \underline{some} \texttt{\{OBJECT\}} have the color \texttt{<mask>} & 27.3 \\
It is known that \underline{few} \texttt{\{OBJECT\}} have the color \texttt{<mask>} & 22.5  \\
It is known that \underline{no} \texttt{\{OBJECT\}} have the color \texttt{<mask>} & 14.0 \\
\midrule
\multicolumn{2}{c}{\textit{paraphrases of the original prompt}} \\
\midrule
It is known that color of most \texttt{\{OBJECT\}} are \texttt{<mask>} & 56.6 \\
It is known that the color of most \texttt{\{OBJECT\}} are \texttt{<mask>} & 59.1 \\
It is common sense that the color of most \texttt{\{OBJECT\}} are  \texttt{<mask>} & \textbf{62.2} \\
It is known that most \texttt{\{OBJECT\}} are \texttt{<mask>} & 49.1 \\
It is known that \texttt{\{OBJECT\}} are \texttt{<mask>} & 44.2 \\
It is common knowledge that most \texttt{\{OBJECT\}} have the color \texttt{<mask>} & 52.0 \\
It is common sense that most \texttt{\{OBJECT\}} have the color \texttt{<mask>} & 55.5 \\
It is commonly known that most \texttt{\{OBJECT\}} have the color \texttt{<mask>} & 53.0 \\
Everybody knows that most \texttt{\{OBJECT\}} have the color \texttt{<mask>} & 54.3 \\
Most people think that \texttt{\{OBJECT\}} have the color \texttt{<mask>} & 53.6 \\
The majority of \texttt{\{OBJECT\}} have the color \texttt{<mask>} & 51.2\\
The vast majority of \texttt{\{OBJECT\}} have the color \texttt{<mask>} & 52.9 \\
Most \texttt{\{OBJECT\}} color \texttt{<mask>} \textit{(ungrammatical)} & 44.1 \\
Common sense most \texttt{\{OBJECT\}} color \texttt{<mask>} \textit{(ungrammatical)} & 43.4 \\
\bottomrule
\end{tabular}
}
    \caption{\gptthree$_{\text{davinci}}$'s 0-shot performance on CoDa across different prompts.}
    \label{tab:prompt}
\end{table*}

\subsection{Discrepancy among Corpora} \label{sec:corpora}
As discussed in Limitations (\Cref{sec:limitations}), we use Google Books and Wikipedia in line with \citet{paik-etal-2021-world} for direct comparison. As can be seen in \Cref{tab:main_results_full}, Google Ngram is better agreeing with human judgment. Moreover, Google Books is much larger than Wikipedia. So, in the main experiments, we use it as an approximation of pretraining corpora. However, it remains unknown how well these sources' ngram distributions align with the real training corpora of LLMs. In future work, there should ideally be more strict control and better access to the pretraining data to draw firmer conclusions. 

\subsection{Have the LLMs seen test data during training?}\label{sec:data_leackage}
It is unlikely that LLMs have seen the test data in its exact form in their pretraining corpora. As the whole web can be used as training data, this is a real risk. However, we think it is unlikely that LLMs have seen CoDa. The CoDa dataset was released on October 2021. GPT-3-davinci-002 was trained with data until June 2021;  GPT-3-curie/babbage/ada-001 were using data until October 2019;\footnote{\href{https://beta.openai.com/docs/models/gpt-3}{beta.openai.com/docs/models/gpt-3}} T5's pretraining corpus C4 was crawled on April 2019. PaLM's precise training data is unknown, but the paper was published after CoDa. However, performance-wise PaLM is not significantly better than GPT-3-davinci-002, which uses training data before the release of CoDa.

\subsection{Error Analysis} \label{sec:error_analysis}
Here we pick the errors made by the models on Single-type questions to understand why or what type of questions they make mistake. Both \gptthree{} and \palm{} achieve above 80\% in this category. We randomly sample 10 errors made by \palm-540b (5-shot) and list them below.

\begin{tcolorbox}
\small
- - - - - - - - - - - - error 1 - - - - - - - - - - - - - \\
query: ... \textbf{most \texttt{mangoes} have the color} \texttt{<mask>} \\
ground truth: \texttt{orange} \\
prediction: \texttt{yellow} \\
- - - - - - - - - - - - error 2 - - - - - - - - - - - - - \\
query: ... \textbf{most \texttt{computer monitors} have the color} \texttt{<mask>} \\
ground truth: \texttt{black} \\
prediction: \texttt{gray} \\
- - - - - - - - - - - - error 3 - - - - - - - - - - - - - \\
query: ... \textbf{most \texttt{sinks} have the color} \texttt{<mask>} \\
ground truth: \texttt{gray} \\
prediction: \texttt{white} \\
- - - - - - - - - - - - error 4 - - - - - - - - - - - - - \\
query: ... \textbf{most \texttt{porcupines} have the color} \texttt{<mask>} \\
ground truth: \texttt{brown} \\
prediction: \texttt{black} \\
- - - - - - - - - - - - error 5 - - - - - - - - - - - - - \\
query: ... \textbf{most \texttt{potatoes} have the color} \texttt{<mask>} \\
ground truth: \texttt{brown} \\
prediction: \texttt{white} \\
- - - - - - - - - - - - error 6 - - - - - - - - - - - - - \\
query: ... \textbf{most \texttt{kangaroos} have the color} \texttt{<mask>} \\
ground truth: \texttt{brown} \\
prediction: \texttt{gray} \\
- - - - - - - - - - - - error 7 - - - - - - - - - - - - - \\
query: ... \textbf{most \texttt{pancakes} have the color} \texttt{<mask>} \\
ground truth: \texttt{brown} \\
prediction: \texttt{yellow} \\
- - - - - - - - - - - - error 8 - - - - - - - - - - - - - \\
query: ... \textbf{most \texttt{scorpions} have the color} \texttt{<mask>} \\
ground truth: \texttt{brown} \\
prediction: \texttt{black} \\
- - - - - - - - - - - - error 9 - - - - - - - - - - - - - \\
query: ... \textbf{most \texttt{coins} have the color} \texttt{<mask>} \\
ground truth: \texttt{gray} \\
prediction: \texttt{yellow} \\
- - - - - - - - - - - - error 10 - - - - - - - - - - - - - \\
query: ... \textbf{most \texttt{picnic baskets} have the color} \texttt{<mask>} \\
ground truth: \texttt{brown} \\
prediction: \texttt{red} \\
\end{tcolorbox}

Most of the ten queries seem to be ambiguous. Black and brown scorpions are both common; the color of a mango might be described as orange or yellow; kitchen sinks are normally gray but bathroom sinks are normally white; old computer monitors are normally gray but newer ones are normally black. The most obvious mistake seems to be on picnic baskets which \palm{} classifies as red. We believe these are included in Single-type questions due to the method used for constructing CoDa. To identify if an object has a single, multiple, or many typical colours, \citet{paik-etal-2021-world} use a clustering algorithm together with manual assignment. However, the threshold of one-versus-many clusters can be hard to decide, and many objects would end up at the boundary. Also, depending on the number of annotators, the presented ground truth may be noisy when compared to the general population.

\section{Few-shot Prompts}\label{sec:few_shot_prompts}

\paragraph{One-shot.} For one-shot, we prepend one randomly selected example from the dataset. The example is constructed by randomly selecting an object from the dataset and then choosing the colour with the highest probability answer from the ground truth. Some of the objects could have multiple reasonable colours (e.g., yellow will be chosen for bananas, even though they can be green or brown).
\begin{tcolorbox}
\small
    \textbf{It is known that most} \{\texttt{OBJECT$_1$}\} \textbf{have the color} \{\texttt{COLOR}$_1$\}; \textbf{most} \{\texttt{OBJECT$_q$}\} \textbf{have the color} \texttt{<mask>}
\end{tcolorbox}

\paragraph{Five-shot.} Similar to one-shot, but we randomly sample five objects from the dataset.

\begin{tcolorbox}
\small
    \textbf{It is known that most} \{\texttt{OBJECT$_1$}\} \textbf{have the color} \{\texttt{COLOR}$_1$\}; \{\texttt{OBJECT$_2$}\} \textbf{have the color} \{\texttt{COLOR}$_2$\}; ...; \{\texttt{OBJECT$_5$}\} \textbf{have the color} \{\texttt{COLOR}$_5$\}; most \{\texttt{OBJECT$_q$}\} \textbf{have the color} \texttt{<mask>}
\end{tcolorbox}

\end{document}

%% file: tex/introduction.tex
\section{Introduction}
Large language models (LLMs) have been compared to hypothetical giant octopi living underwater that are exposed to a lot of language data \citep{bender-koller-2020-climbing}. Such octopi would struggle to understand what actually happens on land as they lack the physical perceptual experience of living there. As such, they may overfit to text-only corpora and thus amplify reporting bias \citep{gordon2013reporting} rather than faithfully reflecting the physical world.

\begin{figure}
    \centering
    \includegraphics[width=0.95\linewidth]{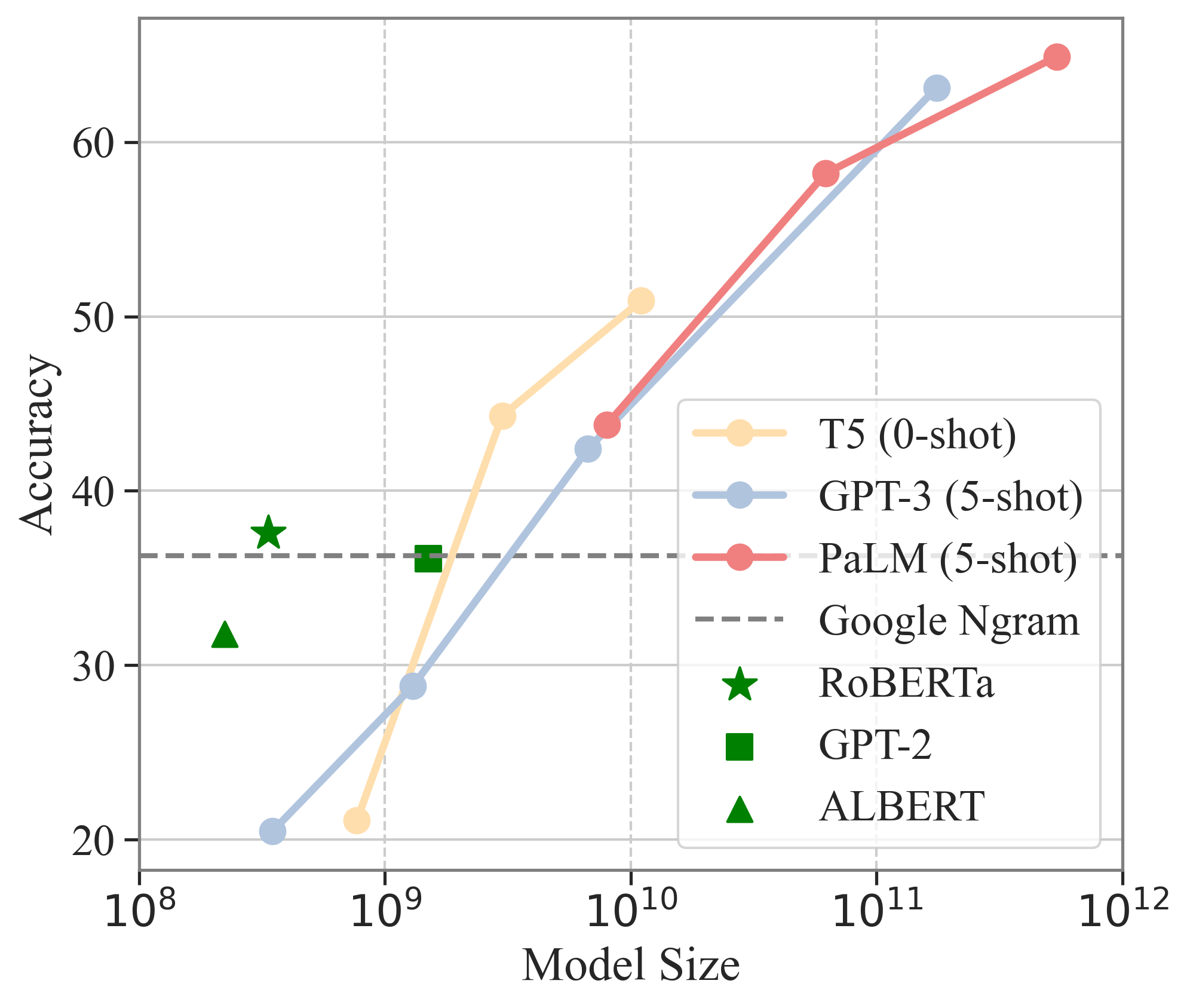}
    \vspace{-0.3cm}
    \caption{On typical colour judgments, large language models (LLMs) greatly outperform small LMs which previously were found to be no better than corpus statistics (Google Ngram). See \Cref{tab:main_results} for full results.}
    \label{fig:main_fig}
    \vspace{-0.5cm}
\end{figure}

In textual corpora, humans do not tend to mention what is commonly known, instead using language to express new information, which is likely less common. For example, when describing the colour of a banana: ``green banana'' has much higher frequency than ``yellow banana'' in the Google Books corpus.\footnote{\href{https://research.tiny.us/google-ngrams-banana}{research.tiny.us/google-ngrams-banana}} It is natural to expect LMs would overfit to such reporting bias since they are trained to memorise such co-occurrence statistics. 
To observe this, we can query widely used pretrained models, such as \roberta$_{\text{Large}}$ \citep{liu2019roberta} with our previous example. Given the prompt ``It is commonly known that most bananas have the color <mask>'', \roberta{} ranks ``green'' the highest.\footnote{\href{https://research.tiny.us/roberta-banana}{research.tiny.us/roberta-banana}}
This agrees with corpus statistics derived from raw text corpus such as the Google Ngram \citep{lin-etal-2012-syntactic} mentioned above.\footnote{The Google Books corpus is an enormous collection of books digitised at Google \citep{michel2011quantitative}. The 2nd edition of the corpus derived by \citet{lin-etal-2012-syntactic} contains >8B books, constituting over 6\% of all books ever published. Google Ngram is a corpus of ngram statistics derived from the Google Books corpus (2nd edition). More details in \Cref{sec:setup}.} \citet{paik-etal-2021-world} test pretrained LMs' perception of colours and confirm that they perform no better than naive co-occurrence statistics extracted from the corpus. In fact, naively using corpus statistics achieves around 40\% accuracy on their proposed colour probing benchmark CoDa while the best LM performs similarly. \citet{zhang-etal-2022-visual} extend the evaluation to a broader range of visual properties, confirming that reporting bias can negatively influence model performance and increasing model size does not help. \citet{shwartz-choi-2020-neural} repeat the reporting bias experiments of \citet{gordon2013reporting} on pretrained LMs and find that LMs overestimate rare events and actions, also amplifying reporting bias.

However, the LMs tested by \citet{paik-etal-2021-world,zhang-etal-2022-visual,shwartz-choi-2020-neural}, i.e., \gpttwo{} \citep{radford2019language}, \bert{} \citep{devlin-etal-2019-bert}, \roberta, and \albert{} \citep{Lan2020ALBERT}, usually have only several hundred million parameters and are of much smaller sizes than LLMs available now. In this work we probe T5 \citep{raffel-etal-2020-exploring}, \gptthree{} \citep{brown2020language}, and \palm{} \citep{chowdhery2022palm} of various sizes, with parameter counts ranging from 770M to 540B. Surprisingly, we find that LLMs almost double the performance of small language models (SLMs) on the typical colour task (\Cref{fig:main_fig}).\footnote{For convenience and consistency, we refer to all models with fewer than 10B parameters as small language models (SLMs) while those with more than 10B parameters as LLMs.} \citet{paik-etal-2021-world} point out that SLMs achieve poor performance on objects that typically only have one colour (such as bananas), possibly due to their true colour being an aspect of common sense and thus not frequently mentioned in the training corpus. We find in contrast that LLMs achieve surprisingly good performance in this category, reaching >80\% accuracy. After plotting accuracy against model size, we observe that scaling up is universally helpful for improving LLMs' performance on the colour probing benchmark (CoDa;  \citealt{paik-etal-2021-world}). Moreover, as LLMs are scaled their correlation to corpus ngram statistics plateau, suggesting that memorising (local) co-occurrence statistics cannot explain their success.\footnote{A careful reader would note here that the models' training data may differ distributionally from Google Ngram. We discuss this more in \Cref{sec:limitations}.}


Our study presents controlled analyses on the colour prediction task as a case study to show that scaling up LLMs could overcome surface-level pattern memorisation (i.e., text reporting bias in our case) and learn physical world common sense at least to some extent. This is an important and surprising finding as it provides a key evidence to counterargue the previous consensus that despite achieving better performance for a range of NLP tasks, larger LMs are more prone to overfitting to corpus statistics and therefore amplifying the reporting bias. Our study points out that this criticism on model scale is misleading as it is not based on the complete picture, and when the model capacity is increased  to a significantly large scale such as \palm-540B and \gptthree$_{\text{davinci}}$, they start to overcome reporting bias and are able to abstract physical common sense from text.

%% file: tex/methods.tex
\section{Method}
To test whether LLMs replicate corpus biases rather than human judgment in the typical colour task, we compare the models' output distributions with the corpora's distribution and the distribution of human judgments. Visual perception provides an ideal testbed as corpus statistics can vary from physical facts; obvious facts are left unspoken. In this case, we focus on the typical colour task, largely following the setup by \citet{paik-etal-2021-world}. Given a query asking the colour of an object, the model must output a distribution over eleven possible colours. We then compare the output distribution to both corpus statistics and average human judgement to examine their respective correlations.

In the following, we explain how we query the LLMs and use their predictions. We test LLMs in three setups: zero-shot, one-shot, and five-shot.

\paragraph{Zero-shot.} We use the following prompt across all models:

\begin{tcolorbox}
\small
    \textbf{It is known that most} \{\texttt{OBJECT$_q$}\} \textbf{have the color} \texttt{<mask>}

\end{tcolorbox}
\noindent where \{\texttt{OBJECT$_q$}\} is replaced with the object's name (in plural form).\footnote{We try other prompts to test LLMs' sensitiveness towards the exact terms used. See \Cref{sec:prompt} for more discussion.} After inputting the prompt, we compute next-token-prediction likelihood for all 11 colours in the CoDA label space and record the log-likelihood scores for all answers as the output distribution of the query:
\begin{equation}
    S(c) = \log P_\Theta (c|\text{prompt})
\end{equation}
where $\Theta$ is LM's parameters; $c$ is the color; ``prompt'' is the input prompt specified earlier. For 0- and 5-shot prompting, the answer scoring scheme remains the same. See \Cref{sec:few_shot_prompts} for details of how few-shot prompts are constructed.


%% file: tex/results.tex


\section{Experimental Setup}\label{sec:setup}

\paragraph{Dataset.} The CoDa dataset contains queries and human judgments of 521 objects.
For each object, CoDa has a human-perceived colour distribution over 11 basic colours in English. The 11 colours were identified by \citet{berlin1969basic} and include black, blue, brown, grey, green, orange, pink, purple, red, white, yellow.
As an example of the dataset, the object ``Carrot'' has the human-perceived scores of black: 0.0, brown: 0.023, orange: 0.797, etc., where the scores over 11 colours sum up to 1.
CoDa contains three types of questions (1) Single (2) Multi and (3) Any. ``Single'' means the object has only one typical colour such as ``Carrot'' which is typically orange. ``Multi'' objects have between two and four typical colours: ``Apple'' is frequently red or green. ``Any'' objects have no fixed set of typical colours, such as ``Shirt'' and ``Car''. By default we report micro-average results across all three types. However, we also discuss the ``Single'' category in detail as it is thought to be especially indicative of reporting bias because such facts are rarely stated in texts.
The statistics of CoDa are listed in \Cref{tab:coda_stats}.\footnote{The original CoDa dataset has a train/validation/test split used for training classifiers to probe embedding-based representations. However, the split was only applied on the embedding model CLIP \citep{radford2021learning} and all other numbers were reported on the full set. To be consistent, we also report performance on the full dataset.} 

\begin{table}[ptbh]
    \centering
    \scalebox{0.75}{
    \begin{tabular}{lccc}
    \toprule
Type & Size & Examples \\
\midrule
Single & 198 & Carrot, Spinach \\
Multi & 208 &  Apple, Street light \\
Any & 115 & Shirt, Car \\
    \bottomrule
    \end{tabular}
    }
\vspace{-0.3em}
\caption{CoDa statistics and examples.}
\label{tab:coda_stats}
\vspace{-1.2em}
\end{table}

\paragraph{Metrics.} We use Acc$_{@1}$, $\rho_{\text{human}}$, $\rho_{\text{ngram}}$. Acc$_{@1}$ measures whether the model gets the most typical colour of an object correct. Other metrics are useful, but less clearly interpretable: $\rho_{\text{human}}$ measures a set of predictions' Spearman's $\rho$ correlation with the distribution of human colour judgments (however, there is low human consensus for some objects and colours). Higher Acc$_{@1}$ is better; higher $\rho_{\text{human}}$ indicates a closer match to human judgments. $\rho_{\text{ngram}}$ measures the models' predictions' correlation with the Google Ngram statistics. Fitting corpus statistics is not necessarily good or bad: we report it to see its relationship with both model size and model performance.

\paragraph{Google Ngram baseline.} Together with queries and human judgments, \citet{paik-etal-2021-world} also provide ngram stats collected from Google Books and Wikipedia to compute the correlation with these corpora. Specifically, they consider all bi- and tri-grams containing a colour followed by an object. A corpus-based baseline is then computing the accuracy/correlation between the total ngram counts of colour-object pairs and the human perceived-scores.
We use Google Ngram as the default baseline as Google Books is much larger than Wikipedia and Google Ngram has better correlation with human judgments than Wikipedia. Wikipedia results are reported in \Cref{sec:comprehensive}.

\paragraph{SLM baselines.} We use the best-performing SLMs from \citet{paik-etal-2021-world} as our baselines, which are \roberta$_{\text{Large}}$, \gpttwo$_{\text{XL}}$, and \albert$_{\text{V2-XXL}}$.
One important difference between \citet{paik-etal-2021-world}'s setup is that they create ten different hand-crafted templates and present the best results
per-object for each model. Our work uses a single template across all models and objects. Thus, we are underestimating LLMs' performance compared to the previously reported SLMs' numbers from \citet{paik-etal-2021-world}. Nonetheless, we see that LLMs outperform SLMs by large margins.\footnote{In \Cref{sec:fair_setup}, we show that SLMs' performance can drop to chance-level using the same zero/few-shot evaluation protocol as LLMs. We also demonstrate that when using different prompts, LLMs such as \gptthree$_{\text{davinci}}$'s 0-shot performance can be improved from 55.5\% to 62.2\% (\Cref{sec:prompt}). However, we uniformly use one single prompt for LLMs to avoid over-optimistic results.}

\paragraph{Compared LLMs and their sizes.}
OpenAI does not disclose the exact size of their text models Ada, Babbage, Curie and Davinci. According to \href{https://blog.eleuther.ai/gpt3-model-sizes}{blog.eleuther.ai/gpt3-model-sizes}, they roughly correspond to  350M, 1.3B, 6.7B, and 175B, which we use as the models' parameter counts. For other models (i.e., T5 and \palm), their number of parameters are made clear in the original papers. We list all compared models' sizes in the second column of \Cref{tab:main_results}.



\begin{table*}[!t]
    \centering
    \scalebox{0.75}{
    \begin{tabular}{lcccccccccc}
    \toprule
    & & \multicolumn{3}{c}{0-shot} &  \multicolumn{3}{c}{1-shot} & \multicolumn{3}{c}{5-shot}   \\
    \cmidrule(lr){3-5}     \cmidrule(lr){6-8}     \cmidrule(lr){9-11}
    Model & Size & Acc$_{@1}$ & $\rho_{\text{human}}$ & $\rho_{\text{ngram}}$  &  Acc$_{@1}$ & $\rho_{\text{human}}$ & $\rho_{\text{ngram}}$ &  Acc$_{@1}$ & $\rho_{\text{human}}$ & $\rho_{\text{ngram}}$ \\
    \midrule
    Google Ngram & - &  36.3 & 44.2 & 100.0 & - & - & - & - & - & - \\
    \midrule 
    \roberta$_{\text{Large}}$$^*$ & 335M & \textbf{37.6} & - & - & - & - & - & - & - & - \\
    \gpttwo$_{\text{XL}}$$^*$ & 1.5B & 36.1 & - & - &  - & - & - & - & - & - \\
    \albert$_{\text{V2-XXL}}$$^*$ & 223M & 31.8 & - & - & - & - & - & - & - & -\\
    \midrule
    T5$_{\text{Large}}$ & 770M & 21.1 & 25.7 & 42.2 & 19.4 & 21.0 & 24.5 & 17.9 & 20.7 & 11.8 \\
    T5$_{\text{XL}}$ & 3B & 44.3 & \textbf{57.4} & 60.3 & 39.0 & 48.8 & 55.2 & 42.4 & 47.8 & 60.3 \\
    T5$_{\text{XXL}}$ & 11B & \textbf{50.9} & 49.5 & 57.5 & \textbf{47.2} & \textbf{54.3} & 55.9 & \textbf{48.0} & \textbf{53.4} & 54.1 \\
    \midrule
    \gptthree$_{\text{ada}}$ & 350M  & 17.9 & 15.7 & 48.8 & 21.3 & 24.5 & 46.0 & 20.5 & 25.4 & 42.2 \\
    \gptthree$_{\text{babbage}}$ & 1.3B & 27.6 & 22.1 & 58.0 & 27.6 & 29.8  & 51.7 & 28.8 & 37.1  & 51.9 \\
    \gptthree$_{\text{curie}}$ & 6.7B & 33.6 & 32.8 & 63.5 & 40.1 & 44.2 & 59.2 & 42.4 & 47.1 & 57.1 \\
    \gptthree$_{\text{davinci}}$ & 175B & \textbf{55.5} & \textbf{43.1} & 65.0 & \textbf{61.8} & \textbf{60.5}  & 61.0 & \textbf{63.1} & \textbf{62.3} & 55.9 \\
    \midrule 
    \palm-8B & 8B & 29.6 & 34.7 & 61.5 &  39.9 & 38.9 & 64.7 & 43.8 & 52.6 & 62.0 \\
    \palm-62B & 62B & 34.2 & 33.5 & 64.4 & 50.1 & 44.8 & 65.3 & 58.2 & 61.9 & 61.1 \\
    \palm-540B & 540B & \textbf{42.6} & \textbf{46.0} & 66.3 & \textbf{63.9} & \textbf{62.5} & 62.5 & \textbf{64.9} & \textbf{66.2} & 60.1\\
    \bottomrule
    \end{tabular}
    }
    \vspace{-0.3em}
    \caption{Results on CoDa (average over all three types). For Acc$_{@1}$ and $\rho_{\text{human}}$ (the higher the better), the best performing models within each model class are \textbf{boldfaced}. The symbol $^*$ denotes numbers from \citet{paik-etal-2021-world}, which uses a more optimistic protocol, aggregating the best per-object performance over 10 hand-crafted prompts. }
    \vspace{-1em}
    \label{tab:main_results}
\end{table*}

\section{Results}\label{sec:results}

\paragraph{Main results (\Cref{tab:main_results}).} We show our main results in \Cref{tab:main_results}. As a general trend, LLMs with >10B parameters all significantly outperform SLMs with <10B parameters, and performance increases monotonically with scale within each model class. While the SLMs do not perform significantly better than Google Ngram (accuracy 36.3\%), LLMs achieve up to 64.9\% (\palm-540B 5-shot). \palm-540B 5-shot also correlates best with human judgments.
For \palm{} and \gptthree, few-shots are much better than 0-shot;\footnote{We observe that \palm{} 0-shot is relatively poor (significantly worse than \gptthree) and its strength is only shown with few-shot. Similar behaviour of \palm{} is also observed on tasks such as Natural Questions \citep{kwiatkowski-etal-2019-natural}. Since this is not the focus of this paper, we leave discovering the cause for future investigation.} while for T5, 0-shot seems to be the best. 

\paragraph{Results on the ``Single'' colour split (\Cref{tab:single_type}).} The ``Single'' split deserves extra attention as it has the highest human consensus and is also considered to be common sense knowledge, implying it is rarely stated in the corpus \citep{paik-etal-2021-world}. While none of the SLM baselines outperform the Ngram baseline on Acc$_{@1}$, the largest \palm{} and \gptthree{} surpass the Ngram baseline by nearly 40\%. Furthermore, the LLMs' predictions correlate significantly more to human judgments.

We also present an error analysis on the ``Single'' split in \Cref{sec:error_analysis}. Out of the ten errors made by \palm-540B, only one is a clear mistake where the model classifies picnic baskets as red. For other nine errors, the error seems to be associated with the ambiguous nature of the questions or the dataset construction process.

\begin{table}[t!]
    \centering
    \scalebox{0.75}{
    \begin{tabular}{llccc}
    \toprule
    & Model &  Acc$_{@1}$ & $\rho_{\text{human}}$ & $\rho_{\text{ngram}}$\\
    \midrule
    \multirow{11}{*}{\rotatebox{90}{-------------------- 0-shot -----------------}} &    Google Ngram & 43.9 & 44.2 & 100.0 \\
    \cmidrule(lr){2-5}   
    & \roberta$_{\text{Large}}$$^*$ & \textbf{42.9} & \textbf{47.8} & -  \\
    & \gpttwo$_{\text{XL}}$$^*$ & 40.4 & 40.3 & - \\ 
    & \albert$_{\text{V2-XXL}}$$^*$ & 34.3 & 43.7 & - \\
     \cmidrule(lr){2-5}   
    & \gptthree$_{\text{ada}}$ & 20.2 & 16.9 & 47.4 \\
    & \gptthree$_{\text{babbage}}$ & 30.8 & 27.4 & 56.0 \\
    & \gptthree$_{\text{curie}}$ & 39.9  & 39.9 & 62.0 \\
    & \gptthree$_{\text{davinci}}$ & \textbf{71.2} & \textbf{50.7} & 62.2 \\
    \cmidrule(lr){2-5}  
    & \palm-8B & 34.8 & 38.2 & 62.1  \\
    & \palm-62B & 44.4 & 34.3 & 64.1 \\
    & \palm-540B & \textbf{53.0} & \textbf{42.2} & 65.6 \\
    \midrule
  \multirow{7}{*}{\rotatebox{90}{-------- 5-shot -----------}} & \gptthree$_{\text{ada}}$ & 19.7 & 21.7 & 42.3 \\
  & \gptthree$_{\text{babbage}}$ & 32.3 &  35.0 & 50.3 \\
  & \gptthree$_{\text{curie}}$ & 53.5 & 47.3 & 55.9  \\
  & \gptthree$_{\text{davinci}}$ & \textbf{82.3} & \textbf{59.9} & 53.3 \\
    \cmidrule(lr){2-5}  
    & \palm-8B & 53.0 & 50.9 & 60.7   \\
    & \palm-62B & 73.2 & 58.5 & 58.5  \\
    & \palm-540B & \textbf{80.8} & \textbf{63.1} & 57.0 \\
    \bottomrule
    \end{tabular}
    }
    \vspace{-0.4em}
    \caption{Results on CoDa (``Single'' type). 1-shot and T5 results (omitted) follow similar trend as \Cref{tab:main_results}.}
    \vspace{-1.5em}
    \label{tab:single_type}
\end{table}

\begin{figure}[t]
    \centering
    \includegraphics[width=0.88\linewidth]{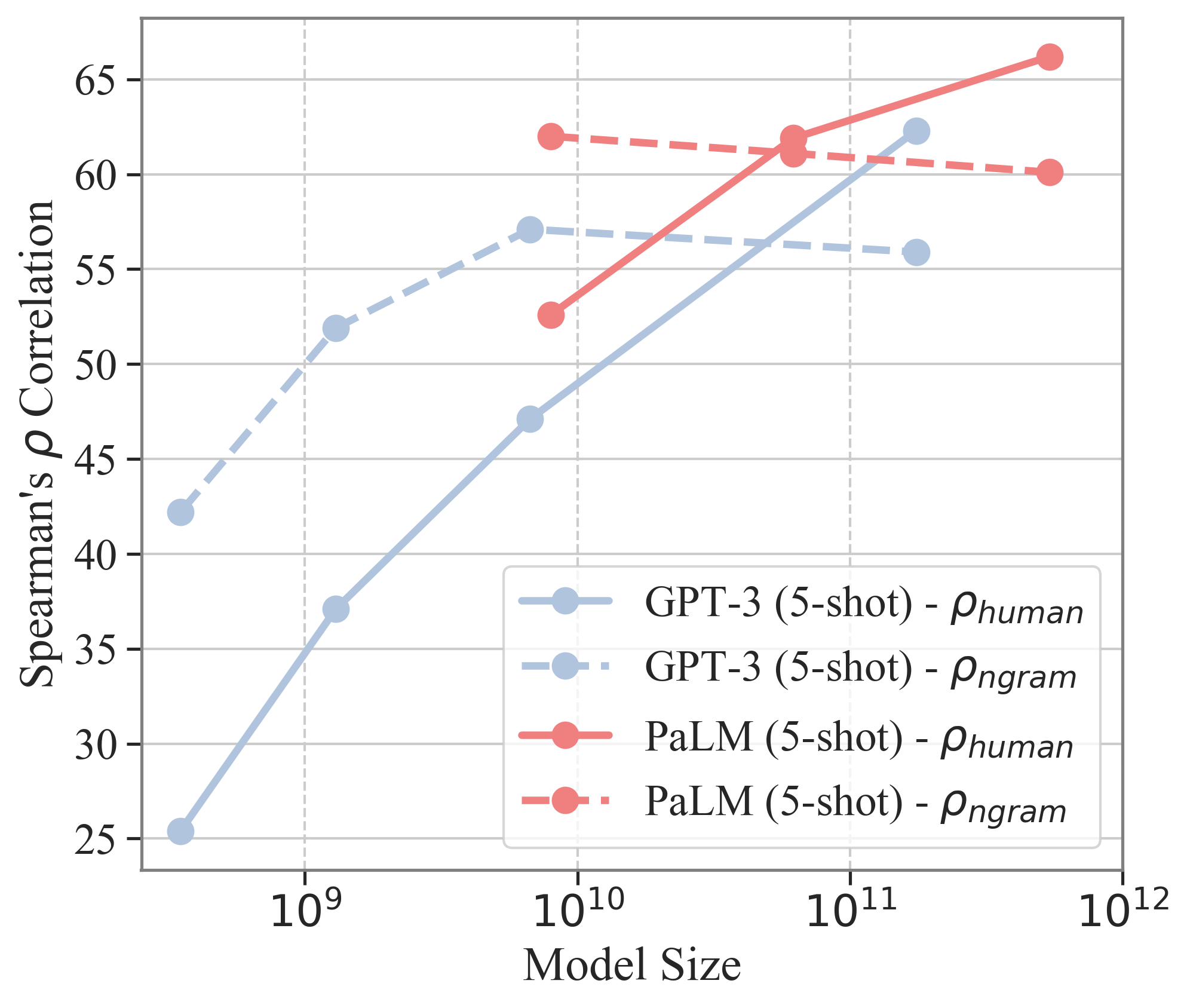}
    \vspace{-0.4em}
    \caption{\gptthree{} and \palm's Spearman's $\rho$ correlation with human judgment and Google Ngram as they are scaled up. These are 5-shots results from \Cref{tab:main_results}.}
    \label{fig:human_vs_corpus}
    \vspace{-0.5em}
\end{figure}

\paragraph{Correlation metrics (\Cref{fig:human_vs_corpus}).} 


For \gptthree, its correlation with corpus ngram statistics ($\rho_{\text{ngram}}$) initially increases but then plateaus and even decreases (on 5-shot: $42.2\rightarrow51.9\rightarrow57.1\rightarrow55.9$).
On \palm, $\rho_{\text{ngram}}$ decreases from the start as model size grows (on 5-shot: $62.0\rightarrow61.0\rightarrow60.1$). 
On both models, $\rho_{\text{ngram}}$ initially is larger than $\rho_{\text{human}}$. However, for model sizes above $10^{11}$ parameters, both models' predictions have $\rho_{\text{human}}>\rho_{\text{ngram}}$.
This suggests that when LMs are small, they can underfit corpus ngrams. When LMs start to be scaled up, they increasingly fit the corpus. However, after a certain model size, additional scale does not lead to more overfitting to corpus statistics. On the contrary, as LLMs' predictions correlate more with human judgment, they also start to decorrelate with corpus statistics.

%% file: tex/limitations.tex
\section{Discussion and Limitations}\label{sec:limitations}

\paragraph{Discrepancy among corpora.}
The corpus statistics we investigate are induced from Google Books and Wikipedia.
They do not necessarily replicate the corpus statistics used for training LLMs. Nonetheless, we do not believe the discrepancy would be big enough to boost LLMs' performance to 80\% on single-type questions. Future work could investigate the original training corpus of LLMs (e.g., C4 for T5).

\paragraph{Is ngram a good reference?} \citet{paik-etal-2021-world,zhang-etal-2022-visual} use the counts of colour occurrences with objects in bi- and tri-grams. However, to fully understand whether LLMs overfit, we also need to consider longer contexts as it is possible that the typical colour of an object is described in longer pieces of text; thus, LLMs performance improvements can be attributed to memorising long-term dependencies better than SLMs. In this case, the ``generalisation'' is only memorising a context that is similar to the prompt. Alternatively, 
 LLMs may learn good representations of the quantifiers, such as ``most'', and the usage of the atypical colours in the text may not co-occur with quantifiers suggesting it is common.
 In future work, we intend to examine whether a similar phenomenon persists when collecting occurrence stats over typical model input lengths and using more fine-grained data that also characterises pre-modifiers such as quantifiers.
 
\paragraph{Comparing within model class for better control of confounders.} Though LLMs today are almost all Transformer-based models with similar autoregressive pretraining objectives, we note that there are caveats preventing us from having a perfect control over design choices on pretraining corpora and specific architectures. 
In terms of pretraining data, within-family models of different sizes generally use the same training data (\gptthree models are however less transparent in this regard). 
However, it is unclear what differences there are across model families. 
In terms of model architectures, T5 is an encoder-decoder model while \gptthree{} and \palm{} are decoder-only models. \palm{} has further modifications on top of the original Transformer architecture such as using SwiGLUE activation \citep{shazeer2020glu} instead of the standard ReLU; using 
RoPE embeddings \citep{su2021roformer} instead of the original relative position embeddings.
As a result, more conclusive findings should be drawn \textit{within} model classes, e.g. comparing \palm-540B with its two smaller versions instead of \gptthree{} models.



\paragraph{Colours live on a spectrum.}
The evidence we obtain does not reflect whether LLMs have a fine-grained and holistic understanding of the nature of colour. That is, colours live on a continuous spectrum. LLMs could have solved CoDa by identifying the mappings between objects and colours but not colour's relative positions on the spectrum. One way to probe this is to examine if LLMs can resolve colour synonyms (e.g., do LLMs know that ``scarlet'' occupies a subspan of the colour red?). However, a rigorous and systematic study of this problem is beyond the scope of this study.

%% file: tex/conclusion.tex
\section{Conclusion and Future Work}\label{sec:conclusion}
In this work, we examine LLMs ability to make typical colour judgments, a simple property of visual common sense. 
Contradicting \citet{paik-etal-2021-world,zhang-etal-2022-visual}, we find that typical colour judgments do not follow an inverse scaling law, and scale is indeed quite critical for high accuracy on the task. While generalising from this task to visual reasoning as a whole is premature, we provide some evidence that larger models of language alone are able to overcome a basic type of reporting bias. Future work will look at a wider range of physical properties \citep{collier2022reality} and more carefully control for the data and model size. We also hope our work opens an avenue for empirically verifying on what level meaning acquisition is possible from a cognitive linguistic perspective \citep{piantasodi2022meaning}.


%% file: tex/ack.tex
\section*{Acknowledgements}
We would like to thank the reviewers and the ACs for their constructive feedback. We would also like to thank Bhuwan Dhingra, William W. Cohen, Tiago Pimentel, Ehsan Shareghi, for the discussions and comments. 
We thank Cory Paik for providing details about \citet{paik-etal-2021-world}.